\pdfoutput=1

\ifdefined\REVIEW
    \documentclass[
        a4paper,
    ]{article}
    \usepackage{lineno, setspace}
    \modulolinenumbers[5]
    \pagewiselinenumbers
\else
    \documentclass[
        a4paper,
        twocolumn
    ]{article}
\fi

\usepackage{IEK10} 

\usepackage{todonotes} \makeatletter
\renewcommand{\todo}[2][]{\tikzexternaldisable\@todo[#1]{#2}\tikzexternalenable}
\makeatother

\def\IEK10{
  Forschungszentrum Jülich GmbH,
  Institute of Energy and Climate Research,
  Energy Systems Engineering (IEK-10),
  Jülich 52425,
  Germany
}
\def\RWTH{
  RWTH Aachen University,
  Aachen 52062,
  Germany
}
\def\KULeuven{
  KU Leuven, 
  Research Center for Operations Management,
  3000 Leuven, 
  Belgium
}
\def\JARA{
  JARA-ENERGY,
  Jülich 52425,
  Germany
}
\def\SVT{
  RWTH Aachen University,
  Process Systems Engineering (AVT.SVT),
  Aachen 52074,
  Germany
}
\def\UQ{
  RWTH Aachen University,
  Chair of Mathematics for Uncertainty Quantification,
  Aachen 52062,
  Germany
}
\def\KAUST{
  King Abdullah University of Science and Technology (KAUST),
  Computer, Electrical, and Mathematical Sciences \& Engineering Division (CEMSE),
  Saudi Arabia
}

\newcommand{\mytitle}{Isometric Manifold Learning for Injective Normalizing Flows}

\newcommand{\affil}{
  \begin{itemize}[leftmargin=3mm, itemsep=0mm]
    \item[$^a$]\IEK10
    \item[$^b$]\SVT
    \item[$^c$]\RWTH
    \item[$^d$]\KULeuven
    \item[$^e$]\JARA
    \item[$^f$]\UQ
    \item[$^g$]\KAUST
  \end{itemize}
}

\def\firstAuthor{Eike Cramer}
\newcommand{\myauthor}{
\firstAuthor$^{a,b}$\orcidlink{0000-0002-6882-5469}, 
Felix Rauh$^{a,c,d}$\orcidlink{0000-0002-9383-4287}, 
Alexander Mitsos$^{e,a,b}$\orcidlink{0000-0003-0335-6566}, 
Ra\'ul Tempone$^{f,g}$\orcidlink{0000-0003-1967-4446}, 
Manuel Dahmen$^{a,*}$\orcidlink{0000-0003-2757-5253} }
\newcommand{\correspondingauthor}{Manuel Dahmen, \IEK10 \newline E-mail: m.dahmen@fz-juelich.de}
\author{\myauthor}

\usepackage[
  colorlinks,
  linkcolor=blue,
  citecolor=blue,
  urlcolor=blue,
  pdftitle={\mytitle},
  pdfauthor={\firstAuthor}
]{hyperref}
\usepackage[capitalise, nameinlink]{cleveref}
\crefname{table}{Tab.}{Tab.}

\usepackage{orcidlink}

\newcommand{\setpgfexternalcounter}[1]{
  \makeatletter \pgfkeysgetvalue{/tikz/external/figure name}\myexternalname
  \expandafter\gdef\csname c@tikzext@no@\myexternalname\endcsname{#1}\makeatother
}

\begin{document}

\ifx\REVIEW\undefined
\twocolumn[
\begin{@twocolumnfalse}
\fi
  \thispagestyle{firststyle}

  \begin{center}
    \begin{large}
      \textbf{\mytitle}
    \end{large} \\
    \myauthor
  \end{center}

  \vspace{0.5cm}

  \begin{footnotesize}
    \affil
  \end{footnotesize}

  \vspace{0.5cm}

To model manifold data using normalizing flows, we employ isometric autoencoders to design embeddings with explicit inverses that do not distort the probability distribution. Using isometries separates manifold learning and density estimation and enables training of both parts to high accuracy. Thus, model selection and tuning are simplified compared to existing injective normalizing flows. Applied to data sets on (approximately) flat manifolds, the combined approach generates high-quality data.  
  \vspace*{5mm}
\ifx\REVIEW\undefined
\end{@twocolumnfalse}
]
\fi

  \newpage

\section{Introduction}

Normalizing flows are deep generative models (DGM) that represent the probability distribution of high-dimensional data sets as a change of variables of a multivariate Gaussian \cite{tabak2010density,tabak2013family}. Using the inverse of this transformation, normalizing flows can compute the probability density functions (PDFs) explicitly, thus enabling training via the statistically consistent and asymptotically efficient \cite{rossi2018mathematical} likelihood maximization \cite{dinh2016realNVP,papamakarios2021normalizing}.

Despite the advantage of likelihood maximization, normalizing flows often result in poor fits of the target probability distributions \cite{brehmer2020flows, behrmann2021understanding}. 
The poor fits often result from the structural setup of the normalizing flow that is ill-posed to describe probability distributions over lower-dimensional manifolds \cite{brehmer2020flows,behrmann2021understanding}.
As the probability distributions of many high-dimensional data sets reside on lower-dimensional manifolds \cite{fefferman2016testing,udell2019big}, the inability of normalizing flows to fit such probability distributions rescinds the advantages gained from the direct likelihood maximization.
Advanced normalizing flow models like Glow \cite{kingma2018glow} and FFJORD \cite{grathwohl2018ffjord} can partially compensate for this issue by relying on modern model architectures.

Approaches to training normalizing flows on manifold data substitute the bijective transformation with an injective transformation, i.e., a pseudo invertible transformation that maps the data to a lower-dimensional Gaussian \cite{gemici2016normalizing, brehmer2020flows}.
Such injective normalizing flows then combine manifold learning with density estimation in a single model and, thus, a simultaneous training approach.
Some of the published approaches assume a dimensionality-reducing map to be known and available a priori \cite{gemici2016normalizing, rezende2020normalizing}. 
Other works use compositions of manifold learning models and normalizing flows that are trained simultaneously, e.g., the $\bm{\mathcal{M}}$-Flow \cite{brehmer2020flows}, Noisy Injective Flows \cite{cunningham2020normalizing}, piecewise injective flows called Trumpets \cite{kothari2021trumpets}, and neural manifold ordinary differential equations \cite{lou2020neural}. 
Approaches without dimensionality reduction use inflation with noise to overcome the manifold structure \cite{horvat2021density,horvat2021denoising,postels2022maniflow}.

Most injective normalizing flows do not support exact PDF tractability and instead require PDF approximations \cite{brehmer2020flows}, approximations of the inverse transformation \cite{cunningham2020normalizing}, or approximations of the Jacobian determinant terms \cite{caterini2021rectangular}. 
Notable exceptions are the conformal embedding flow \cite{ross2021conformal} and the principal manifold flow \cite{cunningham2022principal} that maintain tractable PDFs.
Most injective normalizing flows are designed to perform manifold learning and density estimation using the same transformation. 
Hence, likelihood maximization and manifold learning have to be trained either simultaneously \cite{ross2021conformal} or iteratively \cite{brehmer2020flows}. Both the simultaneous and the iterative training lead to a difficult tuning problem as the loss function has to balance two uneven objectives: The likelihood maximization, which is unbounded \cite{rossi2018mathematical}, and the reconstruction loss of the manifold learning that has a theoretical global optimum of zero. 

In this work, we propose a combination of isometric embeddings and normalizing flows that rescinds the issue of the combined loss function. 
We show theoretically that the PDF described by an injective normalizing flow is invariant to isometric embeddings.
Thus, isometric embeddings allow us to separate the embedding and the normalizing flow training into two distinct problems, and to train both model parts to the highest attainable accuracy.
Meanwhile, the composition still allows for explicit sampling as well as explicit and efficient PDF computation.

This work presents a generalization of our previous work on principal component analysis (PCA)-based dimensionality reduction \cite{cramer2022pricipalcomponentflow} to nonlinear embeddings.
In particular, we use the isometric autoencoder (I-AE) \cite{gropp2020isometric} that enforces the isometry via regularization.
A family of strictly isometric embeddings would be limited to learning flat manifolds \cite{kampler1968flat} and, the regularization approach can balance any trade-offs between the embedding and the learned density \cite{lee2022regularized}. 
Notably, the proposed method does not aim to improve on the expressivity of other injective flows like the $\bm{\mathcal{M}}$-Flow \cite{brehmer2020flows}.
In contrast, our approach simplifies model tuning and training.

The remainder of this paper is organized as follows:
In Section~\ref{sec:methods}, we briefly review the basic principles of normalizing flows, present the general concept of injective normalizing flows, and explicate the difficulties associated with the combination of manifold learning and density estimation.
Then, we show how isometric embeddings allow for a seamless separation of the two learning problems. 
Section~\ref{sec:Iso_Isometric_embeddings} reviews practical designs of isometric embeddings. 
In Section~\ref{sec:NumericalExperiments}, we apply the injective normalizing flows to generate samples of the artificial S-curve data set and the MNIST image data set of handwritten digits. Furthermore, Section~\ref{sec:NumericalExperiments} highlights the limitation of isometric embeddings by the example of a distribution on a spherical surface.
Finally, Section~\ref{sec:Conclusion} concludes our work.   \section{Isometric embedding normalizing flows}\label{sec:methods}
This Section first reviews the basic principles of normalizing flows in Section~\ref{sec:NormFlow}, and then states the general loss function for an injective normalizing flow with combined manifold learning in Section~\ref{sec:InjectiveFlows}. 
Section~\ref{sec:IsometricEmbeddingsGeneral} proposes a separation of manifold learning and density estimation by using isometric embeddings. 
Finally, Section~\ref{sec:iso_limitations_of_iso_embeddings} discusses the limitations of isometric embeddings. 

\subsection{Normalizing flows}\label{sec:NormFlow}
Normalizing flows model the probability distribution of a $D$-dimensional multivariate random variable $X$ through a change of variables of a multivariate standard Gaussian \cite{tabak2013family, papamakarios2021normalizing}.
The change of variables utilizes a diffeomorphism, i.e., a bijective and differentiable transformation, $T(\cdot): \mathbb{R}^D \rightarrow \mathbb{R}^D$. Thus, each point $\mathbf{x}$ in the data probability distribution is uniquely mapped to a point $\mathbf{z}$ in the Gaussian and vice versa. 
The links are given by the forward $T(\cdot)$ and inverse $T^{-1}(\cdot)$ transformation, respectively.
Exploiting the diffeomorphism, the PDF $p_X(\mathbf{x})$ can be calculated explicitly using the change of variables formula \cite{papamakarios2021normalizing}:
\begin{equation}
    p_X(\mathbf{x}) = \phi\left(\mathbf{z}\right) \left| \det \mathbf{J}_{T}(\mathbf{z}) \right|^{-1} \label{Eq:ChangeOfVariablesData}
\end{equation}
Here, $\mathbf{J}_{T}$ is the Jacobian of the transformation $T$, and $\phi(\mathbf{z})$ is the Gaussian PDF. 
Given a data set $\mathcal{X}$ and a trainable diffeomorphism $T_{\bm{\theta}}$ with parameters $\bm{\theta}$, the logarithm of Equation~\eqref{Eq:ChangeOfVariablesData} can be used to train the normalizing flow via direct log-likelihood maximization \cite{papamakarios2021normalizing}.

\subsection{Injective normalizing flows}\label{sec:InjectiveFlows}
Given a $d$-dimensional manifold $\mathcal{M}\subset \mathbb{R}^D$ with $d<D$, let $\psi(\cdot): \mathbb{R}^d \rightarrow \mathcal{M}$ be a decoder function that maps the data from the latent space $\mathbf{z}$ to the data space $\mathbf{x}$ and let $\psi^{-1}(\cdot): \mathbb{R}^D \rightarrow \mathbb{R}^d$ be an encoder function:
\begin{align*}
    \mathbf{x} &= \psi(\mathbf{z}) \\
    \mathbf{z} &= \psi^{-1}(\mathbf{x})
\end{align*}
Then, $\psi^{-1}$ is a pseudo-inverse of $\psi$, i.e., an exact inverse of $\psi$ for $\mathbf{x}\in \mathcal{M}$.
Since using encoder and decoder functions results in non-square Jacobians, the change of variables has to be set up in a generalized form \cite{gemici2016normalizing}:
\begin{align}
    p_X(\mathbf{x}) 
    = \phi(\mathbf{z}) \left| \det \left(\mathbf{J}_{\psi}(\mathbf{z})^T\mathbf{J}_{\psi}(\mathbf{z}) \right)\right|^{-0.5} 
    \label{Eq:ChangeOfVariablesGeneralized}
\end{align}
Note that the generalized change of variables formula in Equation~\eqref{Eq:ChangeOfVariablesGeneralized} is typically prohibitively expensive to compute during training. 

If the injective normalizing flow is used for both density estimation and manifold learning, the reconstruction loss has to be considered in addition to the likelihood maximization. 
The combined loss function then is the expectation of the two parts over the distribution of the random variable $X$ \cite{brehmer2020flows}:
\begin{equation}
    \mathcal{L} = \mathbb{E}_X \left[ -\log~p_X(\mathbf{x}) + \beta~\vert\vert \mathbf{x} - \psi(\psi^{-1}(\mathbf{x})) \vert\vert_2 \right] \label{Eq:CombinedLikelihoodReconstruction}
\end{equation}
In Equation~\eqref{Eq:CombinedLikelihoodReconstruction}, $\vert\vert \cdot\vert\vert_2$ is the Euclidean norm and $\beta$ is a scaling hyperparameter to balance the two loss function components. 
In general, injective normalizing flows require manifold learning and likelihood maximization to be performed either simultaneously \cite{kothari2021trumpets,ross2021conformal} or iteratively \cite{brehmer2020flows}. 
However, this balancing is particularly difficult, as the reconstruction loss $\vert\vert \mathbf{x} - \psi(\psi^{-1}(\mathbf{x})) \vert\vert_2$ is bounded from below by zero, while the likelihood loss $-\log~p_X(\mathbf{x})$ is theoretically unbounded. 
Furthermore, the absolute values of the likelihood vary by orders of magnitude between data sets, which makes tuning the hyperparameter $\beta$ difficult.

\subsection{Isometric embeddings}\label{sec:IsometricEmbeddingsGeneral}
As an alternative to performing both manifold learning and density estimation with a single model, we propose a separation of the two objectives into a composition of two distinct learning problems. 
By leveraging isometric embeddings, we can design dimensionality reduction schemes without distorting the probability distribution \cite{gropp2020isometric}. 
The full transformation $\psi$ then consists of an injective transformation $f(\cdot): \mathbb{R}^d \rightarrow \mathcal{M}$ with pseudo-inverse $g(\cdot): \mathcal{M} \rightarrow \mathbb{R}^d$ and a diffeomorphism $T(\cdot): \mathbb{R}^d \rightarrow \mathbb{R}^d$ in the lower-dimensional latent space:
\begin{align*}
    \mathbf{x} &= \psi(\mathbf{z})= f \circ T(\mathbf{z}) \\
    \mathbf{z} &= \psi^{-1}(\mathbf{x})= T^{-1}\circ g(\mathbf{x})
\end{align*}
Applying the chain rule of differentiation to Equation~\eqref{Eq:ChangeOfVariablesGeneralized}, the change of variables of the composition is given by \cite{brehmer2020flows}:
\begin{equation}
\begin{aligned}
     p_X(\mathbf{x}) =&~\phi(\mathbf{z}) 
     \left| \det \mathbf{J}_T(\mathbf{z})\right|^{-1} \\
     &\left| \det \left(
     \mathbf{J}_f\left(T(\mathbf{z})\right)^T
     \mathbf{J}_f\left(T(\mathbf{z})\right) \right)
     \right|^{-0.5} 
\end{aligned}
 \label{eq:ChangeOfVariablesChainRule}
\end{equation}
Options on how to mitigate the computational cost of Equation~\eqref{eq:ChangeOfVariablesChainRule} typically use approximate expressions for $ \det \left(
     \mathbf{J}_f\left(T(\mathbf{z})\right)^T
     \mathbf{J}_f\left(T(\mathbf{z})\right) \right)$ \cite{brehmer2020flows}.

\begin{figure}
    \centering
    \includegraphics[width=\columnwidth]{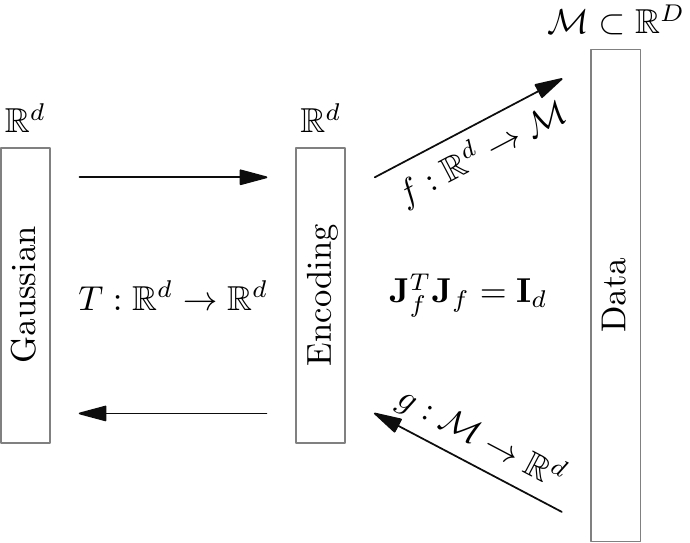}
    \caption{
        Injective normalizing flow designed as a composition of a diffeomorphism $T: \mathbb{R}^d \rightarrow \mathbb{R}^d$ and an isometric embedding $f: \mathbb{R}^d\rightarrow \mathcal{M}$ with pseudoinverse $g:\mathcal{M}\rightarrow \mathbb{R}^d$.
        }
    \label{fig:IsoFlowGeneral}
\end{figure}

In our previous work \cite{cramer2022pricipalcomponentflow}, we used PCA encodings and highlighted that the PDF in Equation~\eqref{eq:ChangeOfVariablesChainRule} is invariant to a PCA encoding.
By definition, Equation~\eqref{eq:ChangeOfVariablesChainRule} is invariant to any other isometric embedding as well.
Mathematically, this can be expressed as
\begin{equation*}
    \mathbf{J}_f(\mathbf{z})^T\mathbf{J}_f(\mathbf{z}) = \mathbf{I}_d \quad \forall\mathbf{z}\in\mathbb{R}^d,
\end{equation*}
where $\mathbf{I}_d$ is the $d$-dimensional identity matrix. 
As the determinant of the identity matrix equals one, the PDF in Equation~\eqref{eq:ChangeOfVariablesChainRule} is invariant to the isometric embedding. 
Therefore, the PDF for any isometric embedding can be simplified to:
\begin{equation}
     p_X(\mathbf{x}) = \phi(\mathbf{z}) \left| \det \mathbf{J}_T(\mathbf{z})\right|^{-1}
\label{Eq:GeneralSimplifiedLikelihoodFunction}
\end{equation}
A visual representation of the two-step composition is shown in Figure~\ref{fig:IsoFlowGeneral}. 

In conclusion, the isometric embedding can be trained independently from the diffeomorphic normalizing flow without any adjustments to the normalizing flow training, which avoids the need to balance the objectives of likelihood maximization and reconstruction loss described in Equation~\eqref{Eq:CombinedLikelihoodReconstruction} and allows us to train both models to high accuracy. 

\subsection{Limitations of isometric embeddings}\label{sec:iso_limitations_of_iso_embeddings}
Restricting the manifold learning problem to strict isometries limits the expressivity of the embeddings. 
The Nash embedding theorem \cite{Nash1956} states that isometric embeddings can encode $d$-dimensional flat Riemannian manifolds embedded in $D$-dimensional ambient space arbitrarily well if $D \geq d + 1$ \cite{gropp2020isometric}.
Thus, the decoder $f$ can be designed to be isometric and the encoder $g$ to be a pseudoinverse of $f$ without loss of information in the embedding, if and only if the considered manifold is flat \cite{kampler1968flat}.
We refer to \cite{lee2022regularized} for a more detailed discussion of the trade-offs between the accuracy and the isometry of the learned embedding for non-flat manifolds.   \section{Isometric embeddings}\label{sec:Iso_Isometric_embeddings}
In this Section, we review two possible isometric embeddings and discuss the selection of the latent dimensionality. 
Note that the use of PCA as affine isometric embedding for normalizing flows was already discussed in our previous work \cite{cramer2022pricipalcomponentflow}.

\subsection{Affine isometric embedding with PCA}\label{sec:AffineIsometryPCA}
For PCA, both the encoder $g_{\text{PCA}}$ and the decoder $f_{\text{PCA}}$ are affine transformations \cite{pearson1901pca}:
\begin{align*}
    \mathbf{x} &= f_{\text{PCA}}(\mathbf{z}) = \mathbf{V}_d \mathbf{z} + \bm{\mu}_\mathcal{X} \\
    \mathbf{z} &= g_{\text{PCA}}(\mathbf{x}) = \mathbf{V}_d^T\left(\mathbf{x} - \bm{\mu}_\mathcal{X}\right)
\end{align*}
Here, $\bm{\mu}_\mathcal{X}$ is the empirical mean of the dataset $\mathcal{X}$ and $\mathbf{V}_d  \in \mathbb{R}^{D\times d}$ are the first $d$ column vectors of the matrix of right singular vectors $\mathbf{V}\in \mathbb{R}^{D\times D}$ that stems from the singular value decomposition (SVD) of the empirical covariance matrix.
Because PCA is an affine transformation, $\mathbf{V}_d$ is the constant Jacobian of the decoding transformation $f_{\text{PCA}}$, hence:
$\mathbf{J}_{f_{\text{PCA}}} = \mathbf{V}_d = \text{const}$. 
By design of the SVD, $\mathbf{V}_d$ has orthonormal column vectors and thus:
\begin{equation*}
    \mathbf{J}_{f_{\text{PCA}}}^T\mathbf{J}_{f_{\text{PCA}}} = \mathbf{V}_d^T\mathbf{V}_d = \mathbf{I}_d
\end{equation*}
  
For PCA, the latent dimensionality $d$ equals the number of considered principal components. We base the selection of $d$ on the explained variance ratio, i.e., the singular values $\sigma_k$ of the empirical covariance matrix normalized by their total sum. The sum over the explained variance ratio of the considered principal components then is the cumulative explained variance (CEV):
\begin{equation*}
    \text{CEV} = \frac{\sum_{k=1}^{d} \sigma_{k}}{\sum_{i=1}^{D} \sigma_{i}} 100\%
\end{equation*}
In practice, we select $d$ such that the CEV is greater or equal to the desired percentage, e.g., 99\%.

\subsection{Nonlinear isometric embeddings with I-AE}\label{sec:NonlinearIsometryIAE}
As a possible nonlinear isometric embedding, we utilize the isometric autoencoder (I-AE) proposed by Gropp et al.~\cite{gropp2020isometric}. Notable alternatives are the regularized autoencoders in \cite{lee2022regularized} and the local conformal autoencoders in \cite{peterfreund2020local}.

The I-AE uses a nonlinear autoencoder and specifically enforces the isometry during training via penalization. Both encoder $g_{\text{I-AE}}$ and decoder $f_{\text{I-AE}}$ are deep neural networks $\textbf{G}(\mathbf{x}; \bm{\theta}_{\textbf{G}})$ and $\textbf{F}(\mathbf{z}; \bm{\theta}_\textbf{F})$, respectively:
\begin{align*}
    \mathbf{x} &=f_{\text{I-AE}} (\mathbf{z})= \textbf{F}(\mathbf{z}; \bm{\theta}_{\textbf{F}}) \\
    \mathbf{z} &=g_{\text{I-AE}}(\mathbf{x}) = \textbf{G}(\mathbf{x}; \bm{\theta}_{\textbf{G}})
\end{align*}
Here, $\bm{\theta}_{\textbf{F}}$ and $\bm{\theta}_{\textbf{G}}$ are the parameters of decoder and encoder, respectively.
To minimize the loss of information, the loss function penalizes the mean-squared-error in the reconstruction:
\begin{equation}
    \mathcal{L}_{\text{AE}}(\bm{\theta}_{\textbf{G}}, \bm{\theta}_{\textbf{F}}) = \mathbb{E}_{X}\left[ \vert \vert \mathbf{x}-\textbf{F}\left(\textbf{G}(\mathbf{x})\right)\vert\vert_2 \right] \label{Eq:IAE_MSE}
\end{equation}
Furthermore, the training has to promote the isometry of the decoder:
\begin{equation}
\mathbf{J}_\textbf{F}(\mathbf{z})^T\mathbf{J}_\textbf{F}(\mathbf{z}) = \mathbf{I}_d \quad \forall\mathbf{z}\in \mathbb{R}^d \label{Eq:IsometryCondition}
\end{equation}
Computing the expression in Equation~\eqref{Eq:IsometryCondition} requires expensive matrix multiplications of the Jacobian and its transpose. 
To avoid these, Gropp et al.~\cite{gropp2020isometric} propose enforcing the isometry condition implicitly by penalizing the change in length of a vector to the surface of the unit ball $\mathbf{u}\in \mathbb{S}^{d-1}=\{\mathbf{y}\in\mathbb{R}^d : \vert\vert \mathbf{y}\vert\vert_2=1 \}$, i.e., a unit vector, after multiplication with the Jacobian:
\begin{equation}
    \mathcal{L}_{\text{iso}}(\bm{\theta}_\textbf{F}) = \mathbb{E}_{Z,U}\left[ \left( \vert\vert \mathbf{J}_\textbf{F}(\mathbf{z}) \mathbf{u}\vert \vert_2 - 1 \right)^2  \right] \label{Eq:L_Iso}
\end{equation}
Here, $\mathbb{E}_{Z,U}$ is the expectation over the latent space $\mathbf{z}\in \mathbb{R}^d$ and the space of all unit vectors $\mathbf{u}\in\mathbb{S}^{d-1}$. 
In addition to the isometry loss of the decoder, the isometry of the encoder (pseudoinverse) has to be promoted as well:
\begin{equation}
    \mathcal{L}_{\text{piso}}(\bm{\theta}_\textbf{G}) = \mathbb{E}_{X,U}\left[ \left( \vert\vert \mathbf{u}^T \mathbf{J}_\textbf{G}(\mathbf{x})\vert \vert_2 - 1 \right)^2  \right] \label{Eq:L_Piso}
\end{equation}
Here, $\mathbb{E}_{X,U}$ is the expectation over the manifold $\mathbf{x}\in \mathcal{M}$ and the space of all unit vectors $\mathbf{u}\in\mathbb{S}^{d-1}$. 
Finally, the three loss functions are combined to a total loss using weights $\lambda_{\text{iso}}$ and $\lambda_{\text{piso}}$ for $\mathcal{L}_{\text{iso}}$ and $\mathcal{L}_{\text{piso}}$, respectively:
\begin{equation}
\begin{aligned}
    \mathcal{L}(\bm{\theta}_\textbf{G}, \bm{\theta}_\textbf{F}) = & \mathcal{L}_{\text{AE}}(\bm{\theta}_\textbf{G}, \bm{\theta}_\textbf{F}) \\
    &+ \lambda_{\text{iso}} \mathcal{L}_{\text{iso}}(\bm{\theta}_\textbf{F}) 
     + \lambda_{\text{piso}} \mathcal{L}_{\text{piso}}(\bm{\theta}_\textbf{G}) 
\end{aligned}
\label{Eq:IAETotalLoss}
\end{equation}

Unlike PCA, the I-AE does not come with a recipe to estimate the latent dimensionality $d$, i.e., $d$ must be determined via empirical studies \cite{gropp2020isometric} or via advanced machine noise inflation techniques \cite{horvat2022intrinsic}. 
For more details on the implementation, selection of the weights $\lambda_{\text{iso}}$ and $\lambda_{\text{piso}}$, and the computation of the loss functions, we refer to the original I-AE publication \cite{gropp2020isometric}.   \section{Numerical experiments}\label{sec:NumericalExperiments}
We apply the proposed combination of isometric embeddings and normalizing flows to learn and sample from the probability distributions of three different data sets. 
Specifically, we analyze the generated samples from standard full-space normalizing flows and isometric embedding flows compared to the real data. 
The three data sets are the synthetic S-Curve data set provided by the machine learning library scikit-learn \cite{scikitlearn}, a two-dimensional Gaussian distribution on a spherical surface,
and the MNIST data set with images of handwritten digits \cite{lecun-mnisthandwrittendigit-2010}. 
For all normalizing flow models, we employ the commonly used RealNVP affine coupling layer architecture \cite{dinh2016realNVP} as the invertible neural network. 
All models are implemented in TensorFlow, version 2.5.0 \cite{tensorflow2015}.

\subsection{S-Curve}
\begin{figure}
    \centering
    \includegraphics[width=\columnwidth]{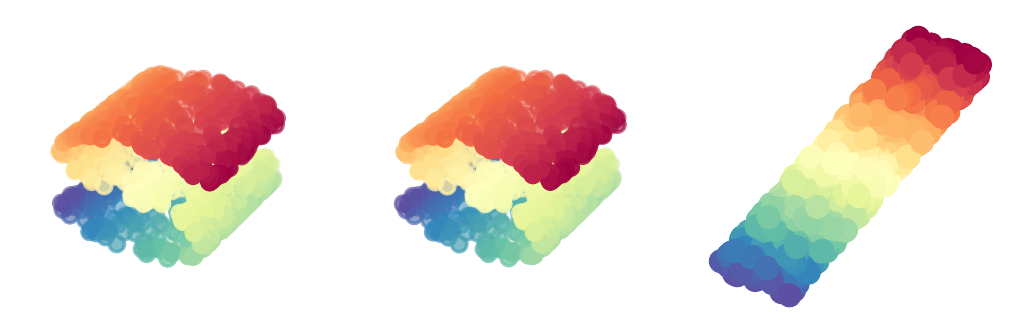}
    \caption{
        Encoding and decoding of three-dimensional S-Curve data (1000 samples) \cite{scikitlearn} using I-AE \cite{gropp2020isometric} with  $\lambda= \lambda_{\textbf{iso}} = \lambda_{\textbf{piso}}=1.0$.
            Test data $\mathbf{x}\in \mathcal{X}_{\text{test}}$ (left);
            reconstruction of test data $f(g(\mathbf{x}))~\forall\mathbf{x}\in\mathcal{X}_{\text{test}}$ (center);
            two-dimensional latent space of I-AE $g(\mathbf{x})~\forall\mathbf{x}\in\mathcal{X}_{\text{test}}$ (right).
        Color coding is based on the position of test data.
        }
    \label{fig:scurveautoencoding}
\end{figure}

The S-Curve data set describes a two-dimensional, nonlinear, flat manifold embedded in three-dimensional Euclidean space (see left of Figure~\ref{fig:scurveautoencoding}). 
Following the training methodology derived in Section~\ref{sec:methods}, we first train the I-AE using Equation~\eqref{Eq:IAETotalLoss} and then train a two-dimensional RealNVP model. 
Figure~\ref{fig:scurveautoencoding} shows the reconstruction (center) and the latent space (right) of the trained I-AE for a test set $\mathcal{X}_{\text{test}}$ of 1000 samples. The results in Figure~\ref{fig:scurveautoencoding} clearly show that the encoding leads to an unfolding of the three-dimensional S-shape to a two-dimensional rectangle and that the full S-Curve is recovered by the decoder. 
Table~\ref{tab:scurveloss} shows the final training and test losses for the total and the three constituents of the loss function. 
\begin{table}
\centering
\caption{
    Final training loss and test loss for I-AE training on the S-Curve data set \cite{scikitlearn}. 
    The constituents of the loss function are defined in Equations~\eqref{Eq:IAETotalLoss}, \eqref{Eq:IAE_MSE}, \eqref{Eq:L_Iso}, and \eqref{Eq:L_Piso}. 
    }
\label{tab:scurveloss}
    \resizebox{\linewidth}{!}{
\begin{tabular}{lllll}
            \hline
            & $\mathcal{L}$                 & $\mathcal{L}_{\text{AE}}$                   & $\mathcal{L}_{\text{iso}}$                   & $\mathcal{L}_{\text{piso}}$                  \\ \hline
Training    & 2.21$\times 10^{-4}$   & 1.55$\times 10^{-4}$   & 4.78$\times 10^{-5}$   & 1.87$\times 10^{-5}$  \\
Test        & 2.11$\times 10^{-4}$   & 1.54$\times 10^{-4}$   & 3.81$\times 10^{-5}$   & 1.89$\times 10^{-5}$  \\ \hline
\end{tabular}
}\end{table}
To verify that the embedding is in fact isometric, we compute the Jacobian determinant of the decoder. For the 1000-sample test set $\mathcal{X}_{\text{test}}$, the expectation of the Jacobian determinant is:
\[  \mathbb{E}_{\mathcal{X}_{\text{test}}} \left[ \left|\det\mathbf{J}_\textbf{F}(\textbf{G}(\mathbf{x}))^T\mathbf{J}_\textbf{F}(\textbf{G}(\mathbf{x}))\right|^{-0.5} \right]=1.0096\] 
Thus, the PDF in Equation~\eqref{Eq:GeneralSimplifiedLikelihoodFunction} only deviates by a factor of 1.0096 from an exact isometry.

\begin{figure}
    \centering
    \includegraphics[width=\columnwidth]{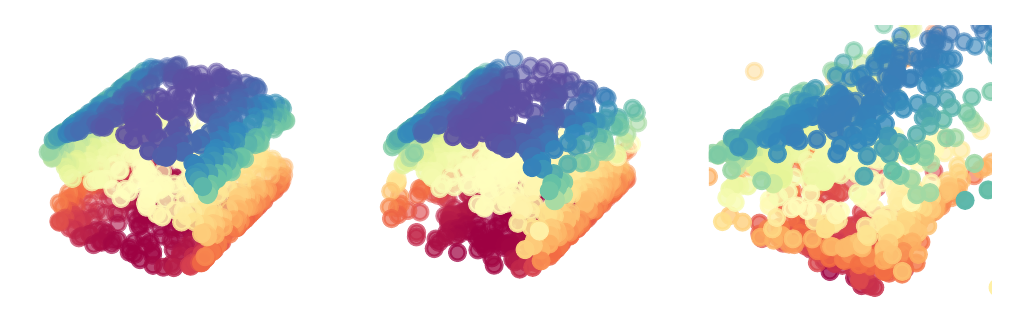}
    \caption{
        1000 generated S-Curve samples \cite{scikitlearn}. 
            Test data $\mathbf{x}\in \mathcal{X}_{\text{test}}$ (left);
            samples from RealNVP in reduced space with I-AE embedding (center); 
            samples from RealNVP in full space (right). 
        The colors are based on the z-axis of the respective data points.  
    }
    \label{fig:scurvesamples}
\end{figure}
Figure~\ref{fig:scurvesamples} shows sampled data of the normalizing flow with I-AE embedding (center) in comparison to a full-space normalizing flow (right). The results show that the manifold shape is correctly recovered by the injective flow, whereas the full-space normalizing flow fails to reconstruct the shape of the S-Curve distribution. Note that we do not show the absolute PDF values of the injective (Equation~\eqref{Eq:GeneralSimplifiedLikelihoodFunction}) and full space normalizing flow (Equation~\eqref{Eq:ChangeOfVariablesData}), as they describe the PDFs of two systems with different dimensionality and are, therefore, not comparable.

The results in Figure~\ref{fig:scurveautoencoding} and Table~\ref{tab:scurveloss} show that the I-AE learns embeddings with very high accuracy. 
Furthermore, the learned embedding is almost isometric, which justifies its usage in an injective normalizing flow as proposed in Section~\ref{sec:methods}. 
This finding is confirmed by the generated data in Figure~\ref{fig:scurvesamples} from the combination of I-AE and RealNVP, which reconstructs the true distribution with only a few outliers.

\subsection{Spherical surface}
We highlight the trade-off between obtaining isometric embeddings and low reconstruction losses via a generated Gaussian distribution on a spherical surface. 
The spherical surface data is sampled from a two-dimensional Gaussian distribution with a random covariance matrix and projected to partially cover a three-dimensional spherical surface. 
The two-dimensional Gaussian is parameterized by a zero mean vector and a randomly sampled covariance matrix:
\begin{equation*}
    \mathbf{z} \sim \mathcal{N}\left(\mathbf{z}; \begin{bmatrix}
0.0\\
0.0
\end{bmatrix}, 
\begin{bmatrix}
0.548 & 0.602 \\
0.602 & 0.544 
\end{bmatrix}
\right)
\end{equation*}
The projection onto the spherical surface is done via stereographic projection:
\begin{equation*}
\begin{bmatrix}
x_1 \\
x_2 \\
x_3
\end{bmatrix} = \frac{1}{1+z_1^2+z_2^2} 
\begin{bmatrix}
2z_1 \\
2z_2 \\
-1+z_1^2+z_2^2
\end{bmatrix}
\end{equation*}
    
\begin{figure}
    \centering
\includegraphics[width=\columnwidth]{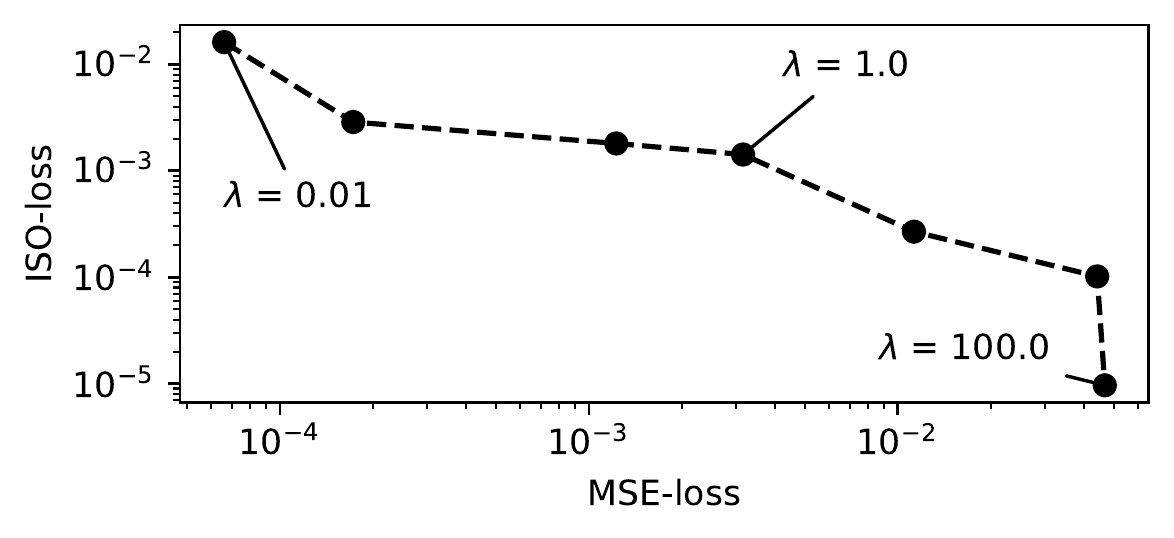}
    \caption{
    Pareto curve of reconstruction loss (MSE-loss, Equation~\eqref{Eq:IAE_MSE}) and isometry loss (ISO-loss, Equation~\eqref{Eq:L_Iso}).
    The parameter $\lambda= \lambda_{\textbf{iso}} = \lambda_{\textbf{piso}}$ is the weighting parameter between reconstruction loss and isometry loss (see Equation~\eqref{Eq:IAETotalLoss}). Dots are validation losses for different values of $\lambda$ and dashed lines are linear interpolations of the Pareto curve for illustration purposes. 
}
    \label{fig:Iso_Limitation_Loss}
\end{figure}
Figure~\ref{fig:Iso_Limitation_Loss} shows the Pareto front of the reconstruction loss (Equation~\eqref{Eq:IAE_MSE}) and the isometry loss (Equation~\eqref{Eq:L_Iso}) for different values of the weighting factor $\lambda = \lambda_\text{iso} = \lambda_\text{piso}$ and a test set $\mathcal{X}_\text{test}$ of 1000 samples.
The Pareto front shows that neither of the two loss functions can be improved without worsening the other indicating a trade-off between the two objectives.

\begin{figure}
    \centering
    \includegraphics[width=\columnwidth]{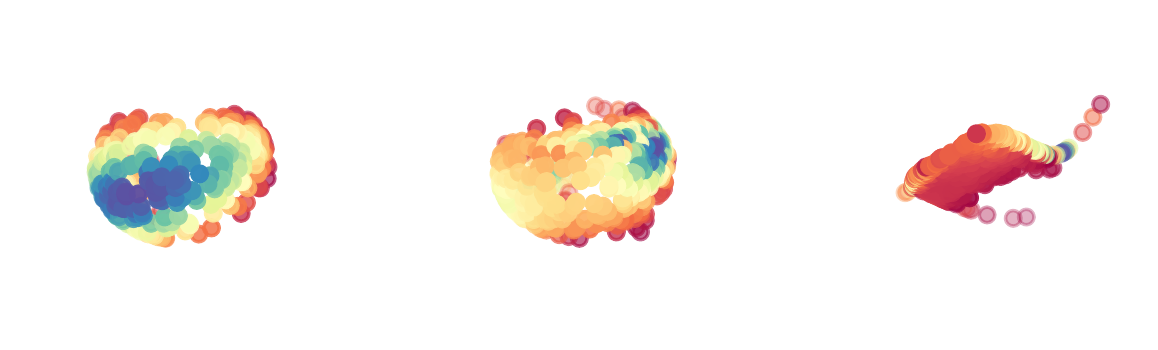}
    \caption{
    Gaussian distribution on a spherical surface. Ground truth (left), I-AE trained with $\lambda = 10^{-3}$ (center), I-AE trained with $\lambda = 10$ (right). The parameter $\lambda= \lambda_{\textbf{iso}} = \lambda_{\textbf{piso}}$ is the weighting factor between reconstruction loss and isometry loss (see Equation~\eqref{Eq:IAETotalLoss}.
    }
    \label{fig:Iso_Limitation_Sphere}
\end{figure}
Figure~\ref{fig:Iso_Limitation_Sphere} shows samples of the true distribution (left), a composition of RealNVP, and an I-AE with low weighting factor ($\lambda=0.01$) on the isometry (center) and with high weighting factor ($\lambda=100$) on the isometry (right), respectively. 
The I-AE with $\lambda=0.01$ learns the shape of the spherical surface well but fails to learn the probability density on the surface. 
The I-AE with $\lambda=100$ fails to learn the surface and, thus, the composition fails to learn the distribution.

\begin{table}
    \centering
    \caption{Average residual of surface equation $R = \vert -1 + \sum_{i=1}^3 x_i^2 \vert$ for 1000 samples from compositions of RealNVP \cite{dinh2016realNVP} and I-AE \cite{gropp2020isometric} for different $\lambda= \lambda_{\textbf{iso}} = \lambda_{\textbf{piso}}$. 
    }
    \label{tab:Surface_test}
\begin{tabular}{llllllll}
        \hline
$\lambda$  & 0.01      & 0.1       & 0.5        & 1          & 5          & 10         & 100       \\ \hline
$R$ & 0.03 & 0.05 & 0.11 & 0.12 & 0.19 & 0.29 & 0.29  \\ \hline
\end{tabular}
\end{table}
Table~\ref{tab:Surface_test} shows the average residuals of the surface equation for sampled data using the isometric manifold normalizing flow with different values of $\lambda$. The trend shows that the representation of the spherical surface becomes worse with increasing $\lambda$. In fact, the average residual increases by an order of magnitude by changing $\lambda$ from 0.01 to 100.

The different extremes in Figure~\ref{fig:Iso_Limitation_Sphere} and the residuals in Table~\ref{tab:Surface_test} show the inability of the I-AE to deal with the curvature of the spherical surface and underline that isometries cannot transform data from a lower-dimensional latent space to a non-flat manifold. 
Thus, the trade-off between representing the true shape of the manifold and finding accurate probability density functions must be considered in applications to real data.

\subsection{MNIST}
Finally, we consider the MNIST data set of handwritten digits \cite{lecun-mnisthandwrittendigit-2010} with $28 \times 28$ pixel resolution. 
The MNIST manifold is nonlinear and not strictly flat \cite{lee2022regularized}. Lee et al.~\cite{lee2022regularized} discuss the trade-off between the isometry and the accuracy of the encoding of the MNIST data set.
For both the I-AE and PCA, we use latent dimensionalities of 16 based on the original I-AE paper \cite{gropp2020isometric}. 
\begin{figure}
    \centering
    \includegraphics[width=\columnwidth]{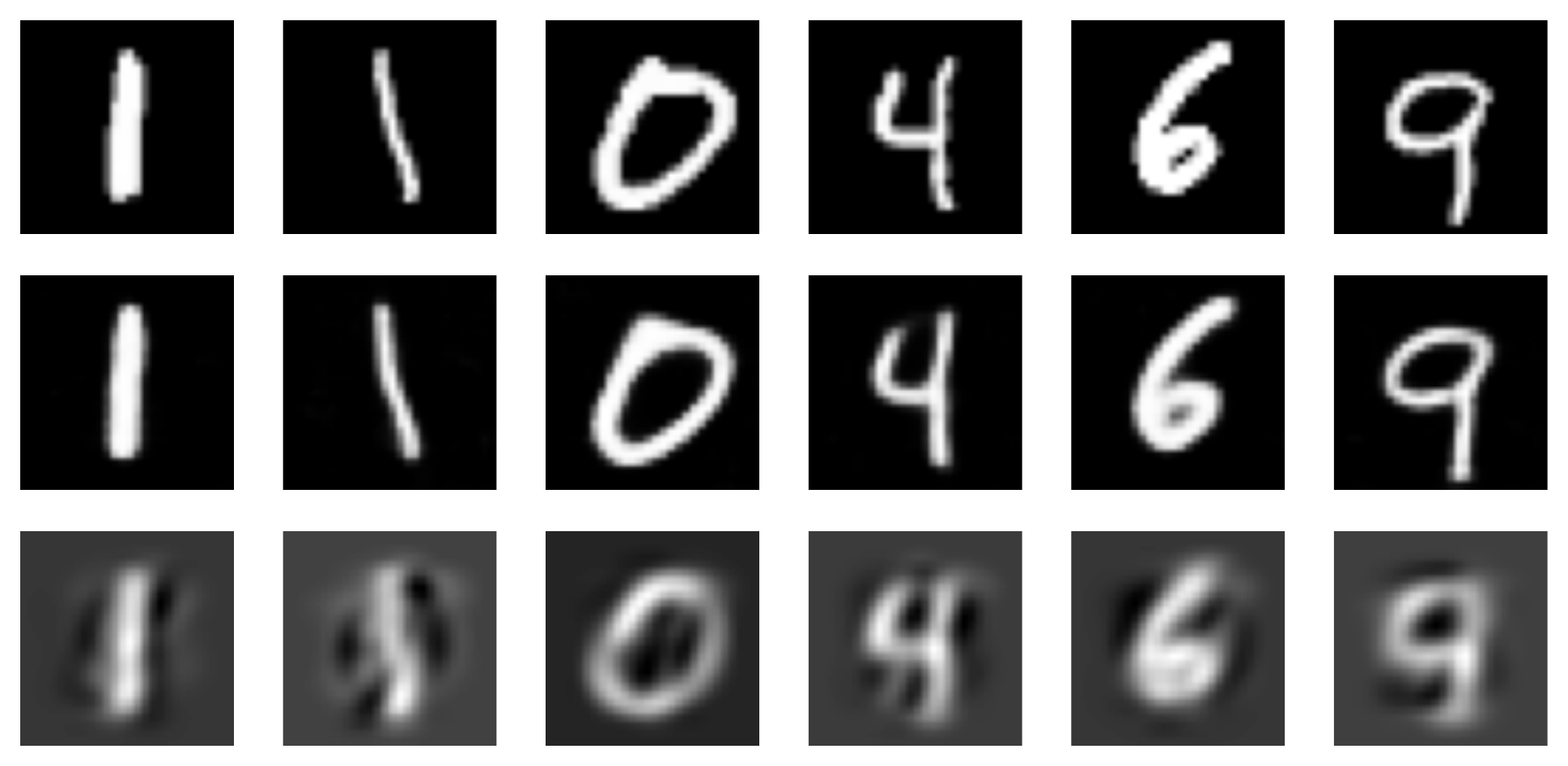}
    \caption{
        I-AE and PCA reconstruction after encoding of MNIST data set of handwritten digits \cite{lecun-mnisthandwrittendigit-2010}. 
        Row 1: Samples from the test set, 
        row 2: Reconstruction after encoding with I-AE, 
        row 3: Reconstruction after encoding with PCA. 
        Latent dimensionalities for I-AE and PCA are 16 based on the original I-AE paper \cite{gropp2020isometric}.
    }
    \label{fig:MNISTIAE}
\end{figure}
Figure~\ref{fig:MNISTIAE} shows a random selection of images from the MNIST test set (first row) as well as the reconstruction of these images after encoding using I-AE (second row) and PCA (third row), respectively. 
For all test images in Figure~\ref{fig:MNISTIAE}, the I-AE is able to preserve the images well. 
The PCA reconstruction shows a loss of features, blurred digits, and a significant amount of noise around the digits. 
\begin{table}
    \centering
    \caption{Reconstruction loss of PCA and I-AE for MNIST test data set \cite{lecun-mnisthandwrittendigit-2010}. The table shows the mean-squared-error (MSE) and the mean-absolute-error (MAE) with data scaled to $[0,1]$.}
    \label{tab:MNIST_Error}
    \begin{tabular}{lrr}
        \hline
            & \multicolumn{1}{l}{I-AE} & \multicolumn{1}{l}{PCA}  \\ \hline
        MSE & 8.28$\times 10^{-3}$ & 2.69$\times 10^{-2}$ \\
        MAE & 3.20$\times 10^{-2}$ & 8.81$\times 10^{-2}$ \\ \hline
\end{tabular}
\end{table}
In addition to the visual comparison in Figure~\ref{fig:MNISTIAE}, Table~\ref{tab:MNIST_Error} lists the mean-squared-error (MSE) and mean-absolute-error (MAE) for both PCA and I-AE on the MNIST test set. With respect to both metrics, the I-AE achieves lower values compared to the PCA. Notably, the MSE for the I-AE encoding is low despite the MNIST manifold being not strictly flat. 
Thus, the nonlinear I-AE shows a better ability to encode the data compared to the affine PCA and overall good performance despite the isometry regularization.

\begin{figure}
    \centering
    \includegraphics[width=\columnwidth]{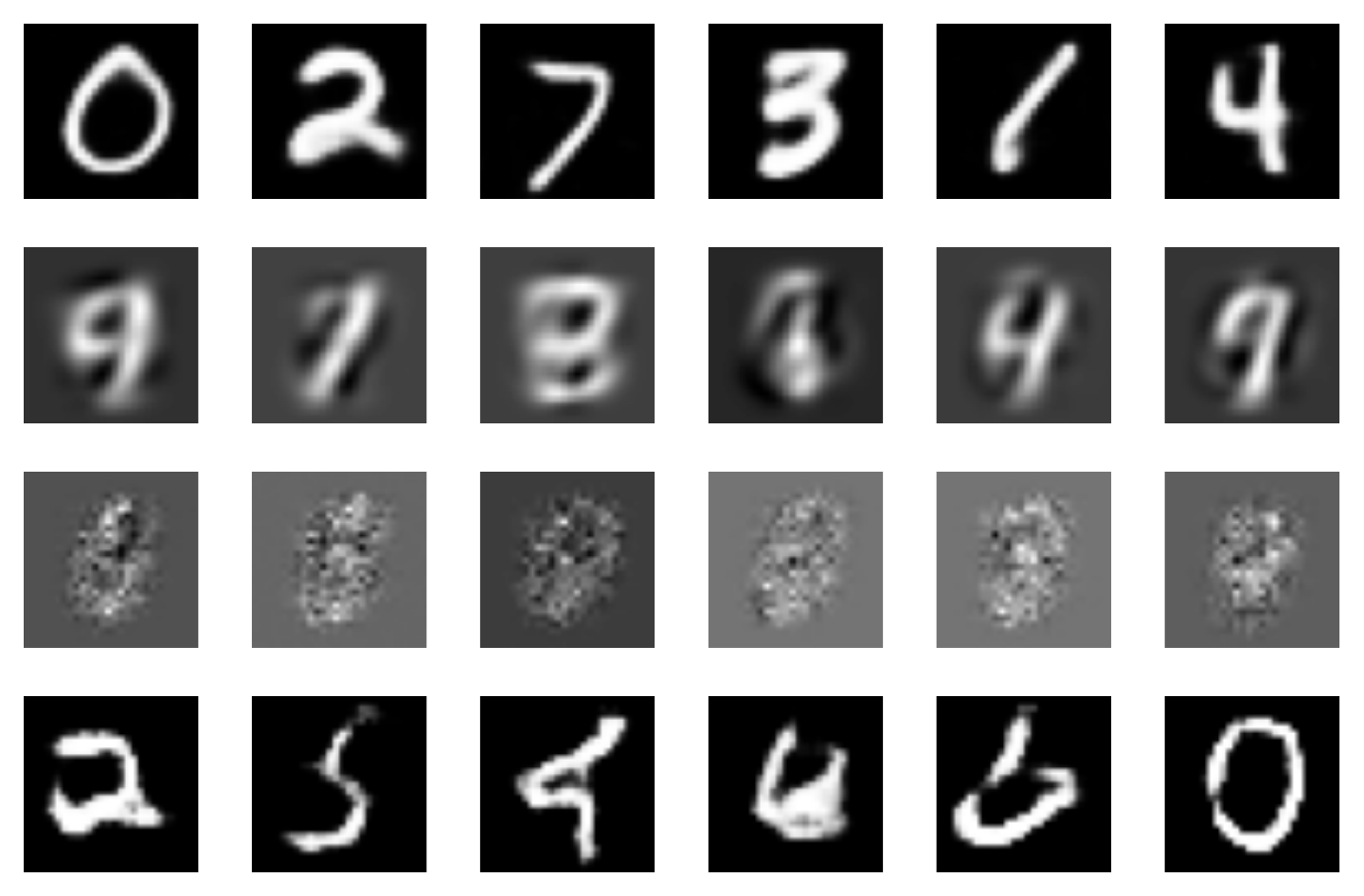}
    \caption{
        Generated images of MNIST data set of handwritten digits \cite{lecun-mnisthandwrittendigit-2010}. 
        Row 1: Random samples of the I-AE and RealNVP composition, 
        row 2: Random samples of the PCA and RealNVP composition, 
        row 3: random samples from a full-space RealNVP,
        row 4: random samples from a Wasserstein GAN \cite{arjovsky2017wasserstein}. 
        Latent dimensionalities for I-AE and PCA are 16.
    }
    \label{fig:MNISTSamples}
\end{figure}
We train different RealNVP models on samples encoded using I-AE and PCA as well as the full images.
Figure~\ref{fig:MNISTSamples} shows samples from the combined isometric embedding normalizing flows in comparison to a full $28\times28$-dimensional normalizing flow. 
For reference, we also include images generated using a Wasserstein-GAN (W-GAN) \cite{arjovsky2017wasserstein}, which is often the method of choice for image generation \cite{bond2021deep}.
Note that the W-GAN was significantly more difficult to train and fine-tune compared to the isometric embedding normalizing flows. Hence, the presented W-GAN results are the best of the 20 trained models. Meanwhile, both isometric normalizing flow compositions converge to the same result in almost every iteration.

The generated images from the composition of I-AE and RealNVP (row 1) appear like handwritten digits from the original MNIST data set. The images generated from the RealNVP with PCA embedding (row 2) resemble handwritten digits to some extent. However, as observed in Figure~\ref{fig:MNISTIAE} the images are blurred, and the digits are difficult to read or ambiguous.
The full-space normalizing flow-generated images (row 3) are very noisy and do not resemble handwritten digits at all, i.e., the full-space RealNVP was unable to learn the probability distribution of the MNIST data set.
The W-GAN-generated images (row 4) show lines with high contrast that resemble digits to some extent. However, these digits are often incomplete or ambiguous. 
Apparently, the discriminator model fails to identify at least some of the images in row 4 as fake. 

In addition to the visual analysis (cf. Figure~\ref{fig:MNISTSamples}), we compute the inception score (IS) \cite{salimans2016improved} and the Fr\'echet inception distance (FID) \cite{heusel2018gans} for each variant of the normalizing flow and the W-GAN. 
Both IS and FID are quantitative measures that analyze both the quality as well as the diversity of the generated images. While the IS considers generated samples only, the FID also includes information from real data.
In general, the IS is a positively oriented score, i.e., high scores indicate high-quality data sets and the FID is a negatively oriented score.
We use a classifier model with $>$99.3\% accuracy. 
As recommended by Salimans et al.~\cite{salimans2016improved}, we generated 50000 samples from each normalizing flow variant to compute both IS and FID. For the IS, we batch the data set and compute the empirical mean and standard deviation. 
Table~\ref{tab:IS} lists IS and FID for the generated image data sets.

\begin{table}
\centering
\caption{
    Inception score (IS) \cite{salimans2016improved} and Fr\'echet inception distance (FID) \cite{heusel2018gans} for MNIST test set $\mathcal{X}_{\text{Test}}$ and generated images from I-AE and PCA compositions with RealNVP, full-space RealNVP (FSNF), and Wasserstein-GAN (W-GAN). 
    The number of generated images is 50000, and the mean and standard deviation for the IS are computed over 10 batches. A high IS and low FID indicate good scores.
    }
\label{tab:IS}
\begin{tabular}{lrr}
    \hline
             & \multicolumn{1}{l}{IS}  & \multicolumn{1}{l}{FID}     \\ \hline
    $\mathcal{X}_{\text{Test}}$ & 11.39   $\pm$ 0.39      & -    \\ 
    I-AE     & 12.10   $\pm$ 0.35      & 46.60   \\
    PCA      & 7.04    $\pm$ 0.25      & 724.51  \\
    FSNF     & 2.55    $\pm$ 0.08      & 1011.60 \\
    W-GAN    & 12.15   $\pm$ 0.51      & 95.10  \\ \hline
    \end{tabular}\end{table}

The results in Table~\ref{tab:IS} highlight that the image data set generated using the I-AE embedding scores as high as the test set with real images with respect to the IS and also shows the lowest FID of all models. The W-GAN scores an IS as good as the I-AE but shows a higher FID than the I-AE. 
The results suggest that the W-GAN images are diverse and can be classified well as indicated by the IS, but do not resemble the actual data as well as the I-AE normalizing flow as highlighted by the FID. 
The IS also correctly indicates a lower quality of the images generated using the PCA embedding as well as the exceptionally low quality of the FSNF-generated images.
Thus, the quantitative results in Table~\ref{tab:IS} confirm our conclusions from the visual assessment in Figure~\ref{fig:MNISTSamples}. 
Note that despite the isometry regularization, the FID score for the I-AE images is competitive with that of, e.g., Ross~and~Cresswell~\cite{ross2021conformal}, who report an FID score of 38.5. 

In conclusion, the isometric dimensionality reduction vastly increases the quality of the generated images and outperforms the W-GAN reference despite the MNIST manifold not being strictly flat.
In fact, the I-AE learns highly accurate encodings of the MNIST data set despite a latent dimensionality as low as 16. 
Our analysis of Figure~\ref{fig:MNISTSamples} and Table~\ref{tab:IS}, thus, shows that the composition of I-AE with RealNVP can build highly descriptive and accurate generative models for data that resides on nonlinear manifolds. 
As expected, the affine PCA shows a distinctively lower quality, which, however, is still far better than the full-space normalizing flow images, which are completely unusable.

   \section{Conclusion}\label{sec:Conclusion}
This work proposes a two-parted approach that separates injective normalizing flows into an isometric manifold-learning part and a lower-dimensional density estimation part.
The probability density function (PDF) described by the change of variables formula is invariant to isometries, which leads to two major benefits in training: 
First, no expensive Jacobian determinant computations result from the embedding. 
Second, embedding functions and normalizing flow models can be trained separately, which avoids the difficult balancing of the unbounded likelihood maximization and the bounded reconstruction loss. 
Thus, both model parts can be trained to the highest attainable accuracy. 

We apply the isometric embeddings to the artificial S-Curve data set, a projection of a Gaussian to a spherical surface, and the MNIST data set with images of handwritten digits. 
For the S-Curve and the MNIST data sets, the isometric embeddings excel at modeling the probability distributions and sample highly realistic data compared to the standard full-space normalizing flow. 
Meanwhile, the spherical surface data set highlights the limitations of isometric embeddings, which are unable to describe curved manifolds. 
The high-quality MNIST samples highlight that the composition of I-AE and RealNVP performs well, despite the trade-off between a strict isometry and a minimal reconstruction loss.
In practice, the complexity of the embedding should be increased progressively, i.e., starting with PCA and progressively transitioning to higher complexity if the embedding performs poorly. 

In conclusion, the proposed combination of isometric embeddings and normalizing flows presents an effective approach to handling manifold data and takes full advantage of the direct log-likelihood maximization and explicit sampling of normalizing flows. 
For established injective normalizing flow models, the typical combined learning approach is difficult to tune. Thus, the separation of embedding and density estimation is of great benefit as both model parts can be trained to high accuracy without trade-offs between the two loss functions. In short, the proposed separation makes training the injective normalizing flow easy.  

\section*{Acknowledgements}
\noindent This work was performed as part of the Helmholtz School for Data Science in Life, Earth and Energy (HDS-LEE) and received funding from the Helmholtz Association of German Research Centres.
This publication is based upon work supported by the King Abdullah University of Science and Technology (KAUST) Office of Sponsored Research (OSR) under Award No. OSR-2019-CRG8-4033, and the Alexander von Humboldt Foundation.
  \section*{Nomenclature}
\noindent\begin{tabularx}{\columnwidth}{Xl}
    Description                     & Symbol   \\ 
    \hline
    Latent dimension                & $d$ \\
    Data dimensionalty              & $D$ \\
    Decoder                         & $\mathbf{f}(\mathbf{z})$ \\
    Encoder                         & $\mathbf{g}(\mathbf{x})$ \\
    Jacobian                        & $\mathbf{J}$ \\
Data manifold                   & $\mathcal{M}\subset \mathbb{R}^D$\\
    Probability density function    & $p_X(\mathbf{x})$ \\
    Surface of $d$-dimensional unit ball & $\mathbb{S}^{d-1}$ \\
    Diffeomorphism                  & $T(\cdot)$ \\
    Samples of multivariate random variables & $\mathbf{x}$,$\mathbf{z}$ \\ 
    Random variable                 & $X$ \\
    Data set                        & $\mathcal{X}$ \\
    Sample mean vector of $\mathcal{X}$ & $\bm{\mu}_\mathcal{X}$  \\
    PDF of standard Gaussian        & $\phi(\mathbf{z})$ \\
    Parameter vector                & $\bm{\theta}$ \\
\end{tabularx}

  \appendix

\bibliographystyle{unsrt}
  \renewcommand{\refname}{Bibliography}
  \bibliography{extracted.bib}

\end{document}